\documentclass[11pt]{article}

\usepackage[final]{acl}
\usepackage{times}
\usepackage{latexsym}

\usepackage[T1]{fontenc}
\usepackage[utf8]{inputenc}

\usepackage{microtype}

\usepackage{inconsolata}

\usepackage{graphicx}

\usepackage{pifont}
\usepackage{multirow}
\usepackage{float}
\usepackage[utf8]{inputenc} % allow utf-8 input
\usepackage[T1]{fontenc}    % use 8-bit T1 fonts
\usepackage{url}            % simple URL typesetting
\usepackage{booktabs}       % professional-quality tables
\usepackage{amsfonts}       % blackboard math symbols
\usepackage{nicefrac}       % compact symbols for 1/2, etc.
\usepackage{microtype}      % microtypography
\usepackage{xcolor}         % colors
\usepackage{pifont}
\usepackage{multirow}
\usepackage{amsmath}
\usepackage{enumitem}
\usepackage{tcolorbox}
\usepackage{colortbl}
\usepackage{ulem}
\usepackage{wrapfig}
\usepackage{subcaption}
\usepackage{tabularx}

\definecolor{lightgray}{gray}{0.9}

\title{BusterX++: Towards Unified Cross-Modal AI-Generated Content Detection and Explanation with MLLM}

\author{
    Haiquan Wen$^{1}$ \quad Tianxiao Li$^{1}$ \quad Zhenglin Huang$^{1}$ \quad Yiwei He$^{1}$ \quad Guangliang Cheng$^{1}$\textsuperscript{$\dagger$} \\
    {$^1$University of Liverpool} \\
    {\normalsize \textsuperscript{$\dagger$} Corresponding author. E-mail: guangliang.cheng@liverpool.ac.uk} \\
    {\normalsize Project Page: \url{https://github.com/l8cv/BusterX}} \\
}

\begin{document}
\maketitle

\begin{abstract}
The rapid advancement of generative AI has substantially improved image and video synthesis, amplifying the risk of multimodal visual misinformation. Recent MLLMs have shown promise for transparent AI-generated content detection through reasoning and explanation, yet existing approaches largely treat image and video forensics as isolated tasks, leaving cross-modal synergies underexplored. 
To address this, we present \textbf{BusterX++}, a unified MLLM for joint image and video detection with interpretable reasoning. We also introduce \textbf{GenBuster-Bench++}, a curated benchmark containing balanced image and video samples spanning recent generation models and diverse real-world scenarios, with synthetic images and videos screened using the same human difficulty criterion. 
Using this controlled setting, we revisit the widely adopted SFT+RL post-training paradigm. Notably, in the unified image-video setting, our results demonstrate that a single-stage pure RL strategy driven primarily by sparse outcome rewards outperforms a strong SFT+RL baseline. We observe that single-stage pure RL maintains higher policy entropy than SFT-initialized RL throughout training. This observation suggests a possible role for exploration in the performance gains observed under unified image-video training.
Extensive experiments demonstrate that BusterX++ achieves state-of-the-art performance, highlighting the powerful potential of RL for unified cross-modal visual reasoning.
\end{abstract}

\section{Introduction}
\label{sec:intro}

The rapid expansion of AI-generated media has escalated the threat of visual misinformation, driving a paradigm shift from traditional binary classifiers to MLLMs that offer reasoning-driven, transparent forensics \cite{zeng-etal-2025-exploring, chang2024antifakepromptprompttunedvisionlanguagemodels, xu2024fakeshield, wen2025busterxmllmpoweredaigeneratedvideo}. While recent frameworks unify image and video detection under one architecture \cite{jiang2025ivy}, a fundamental question remains unaddressed: \textit{Do these distinct modalities exhibit mutual interference or constructive synergy within a shared representation space?}

To investigate cross-modal interaction under more controlled evaluation conditions, evaluating models on fragmented or unbalanced datasets is insufficient. To provide a controlled testbed, we first introduce \textbf{GenBuster-Bench++}, a meticulously curated unified benchmark of AI-generated images and videos. GenBuster-Bench++ contains balanced image and video samples spanning recent generation models and diverse real-world scenarios. Synthetic images and videos are screened using the same human difficulty criterion, providing a unified testbed for evaluating cross-modal capabilities.

Leveraging this calibrated environment, we perform comprehensive investigations that revisit the common assumption of the SFT+RL paradigm. Conventionally, SFT on high-quality annotations is often regarded as a beneficial initialization before RL. To construct a strong baseline reflecting this standard, we fine-tuned our MLLM using 10,000 high-quality reasoning rationales distilled from Qwen3.5-397B. Notably, our empirical results demonstrate that a single-stage, pure RL strategy driven strictly by simple rewards without any SFT warm-up or complex reward shaping consistently yields competitive or superior performance compared to the strong SFT+RL pipeline.

To better understand this behavior, we analyze training dynamics and observe lower policy entropy under SFT initialization. Our single-stage pure RL paradigm sustains higher policy entropy throughout training, suggesting a possible role for exploration in the observed performance difference. Our unified training experiments also suggest \textit{cross-modal capability transfer}. Specifically, our findings suggest that fine-grained texture and sharp boundary priors native to high-resolution images can transfer to mitigate frame-level inconsistencies in synthetic videos. Conversely, the dynamic environmental variations from video sequences can enrich the model's robustness when auditing static images.

Informed by these findings, we present \textbf{BusterX++}, a unified MLLM for cross-modal AI-generated content detection and explanation. BusterX++ adopts a streamlined pure RL training pipeline and is designed to exploit complementary supervision across images and videos. Overall, our work suggests that unified visual forensics may benefit not only from architectural unification, but also from training strategies that encourage transferable reasoning across modalities.

\section{Related Work}
\label{sec:related}

\begin{figure*}[ht]
    \centering
    \includegraphics[width=1\linewidth]{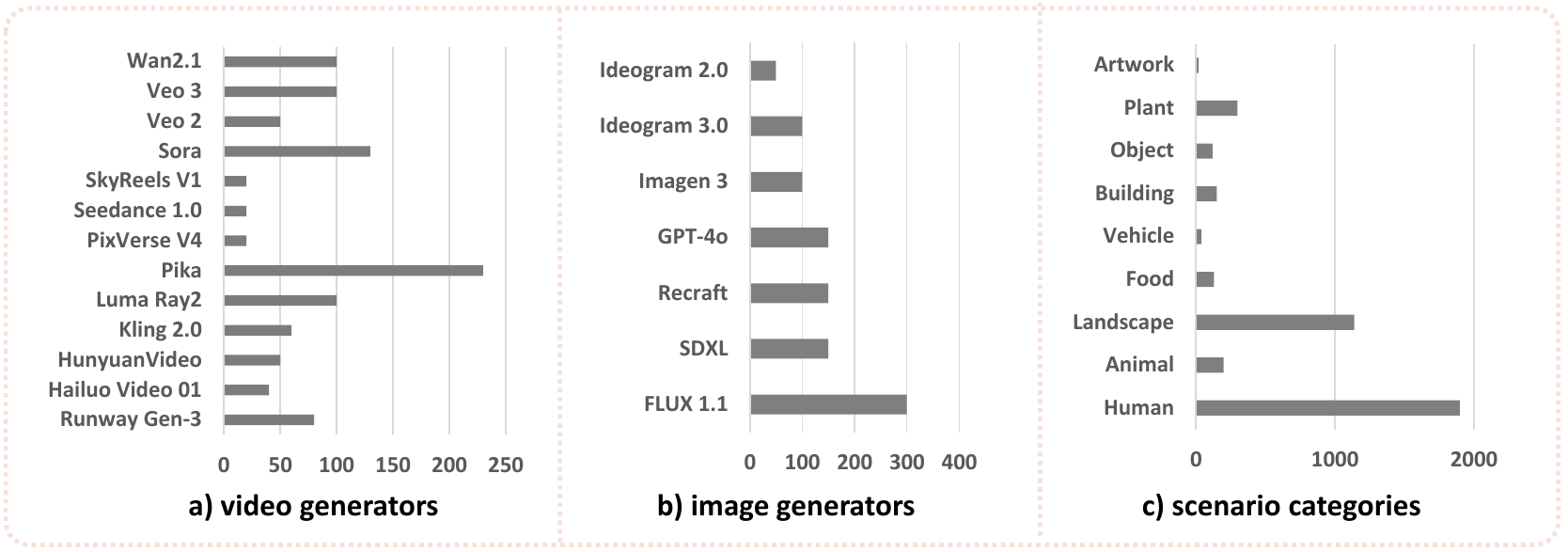}
    \caption{Overview of GenBuster-Bench++. a) GenBuster-Bench++ incorporates video samples generated by 13 SOTA video generators. b) It also includes image samples from 7 SOTA image generators. c) It covers diverse real-world scenarios, including 9 scenario categories and each sample undergoes a rigorous filtering process.}
    \label{fig:benchmark_overview}
\end{figure*}

\subsection{AI-Generated Content Benchmark}
The landscape of AI-generated content detection benchmarks has evolved to reflect the growing complexity of synthetic media. The early benchmarks \cite{dang2020ddf, he2021forgerynet, rossler2019faceforensicspp, khalid2021fakeavceleb, zi2020wilddeepfake, Kwon2021Kodf, xu2024identity} focused mainly on GAN-generated facial forgeries, consistent with the initial focus of deepfake technology on the modification of human identities. With the development of more sophisticated generative models, the focus gradually shifted toward generating diverse and realistic content. In the image domain, benchmarks such as GenImage \cite{zhu2023genimagemillionscalebenchmarkdetecting}, DMImage \cite{corvi2022detectionsyntheticimagesgenerated}, TrueFake \cite{dellanna2025truefakerealworldcase}, MMTD-SET \cite{xu2024fakeshield} have significantly expanded the scope of research. Similarly, the video domain has witnessed substantial progress with benchmarks like GVF \cite{ma2024decof}, GenVideo \cite{chen2024demamba}, GenVidBench \cite{ni2025genvidbench}, GenBuster-200K \cite{wen2025busterxmllmpoweredaigeneratedvideo}. These benchmarks leverage latest video generators to produce high-quality synthetic content that closely mimics real-world scenarios. 
Despite significant advancements, existing benchmarks have notable limitations. Most are confined to single-modality data, restricting their utility in assessing unified capabilities. Additionally, most of them lack fine-grained human curation, resulting in inconsistent quality. To address these gaps, GenBuster-Bench++ integrates both image and video modalities, leveraging two-stage filtering to ensure sample quality, content diversity, and real-world relevance.

\subsection{AI-Generated Content Detection Method}
Traditional AI-generated content detection methods are primarily non-MLLM \cite{afchar2018mesonet, tan2020efficientnetrethinkingmodelscaling, duan2024test, rossler2019faceforensicspp, tan2023learning, bi2023detectinggeneratedimagesreal,  zheng2021exploring, chen2024demamba, wang2020cnngeneratedimagessurprisinglyeasy, ojha2023towards}, focusing on binary classification tasks. These methods achieved reasonable accuracy on in-domain data but exhibited limited generalization to unseen generative techniques and lacked interpretability. Recent advancements introduced MLLM-based methods \cite{chang2024antifakepromptprompttunedvisionlanguagemodels, xu2024fakeshield, huang2025sidasocialmediaimage, chen2025x2dfdframeworkexplainableextendable, wen2025spotfakelargemultimodal, kang2025legionlearninggroundexplain, li2025fakescopelargemultimodalexpert, peng2025mllmenhancedfaceforgerydetection, zhou2025aigiholmesexplainablegeneralizableaigenerated, huang2025sofakebenchmarkingexplainingsocial, song2024learning} in both AI-generated image and video detection fields, which enhanced detection transparency.
However, these approaches remained confined to single-modality inputs, failing to fully leverage the unified capabilities of MLLMs. To address these limitations, we propose BusterX++, which demonstrates superior generalization and adaptability in unified detection and explanation.

% \subsection{Multimodal Large Language Model}
% Recent advancements in MLLM have emphasized enhancing reasoning capabilities. Commercial models like GPT-4o \cite{openai2024gpt4ocard}, Gemini 2 \cite{team2023gemini}, Claude 4 \cite{claude}, and Kimi-k1.5 \cite{kimiteam2025kimik15scalingreinforcement} have demonstrated remarkable capabilities. On the open-source side, models such as Qwen-VL \cite{bai2025qwen25vltechnicalreport} and InternVL \cite{zhu2025internvl3exploringadvancedtraining} stand out. These models allocate additional computational resources to complex reasoning tasks, pushing the limits of existing benchmarks. The community has even introduced extremely challenging benchmarks like Humanity's Last Exam \cite{phan2025humanitysexam} to benchmark the models' limitation. BusterX++ further explores the reasoning capabilities of MLLM in the domain of AI-generated content detection.

\section{Benchmark}
\label{sec:benchmark}

\begin{figure}[t]
    \centering
    \includegraphics[width=1\linewidth]{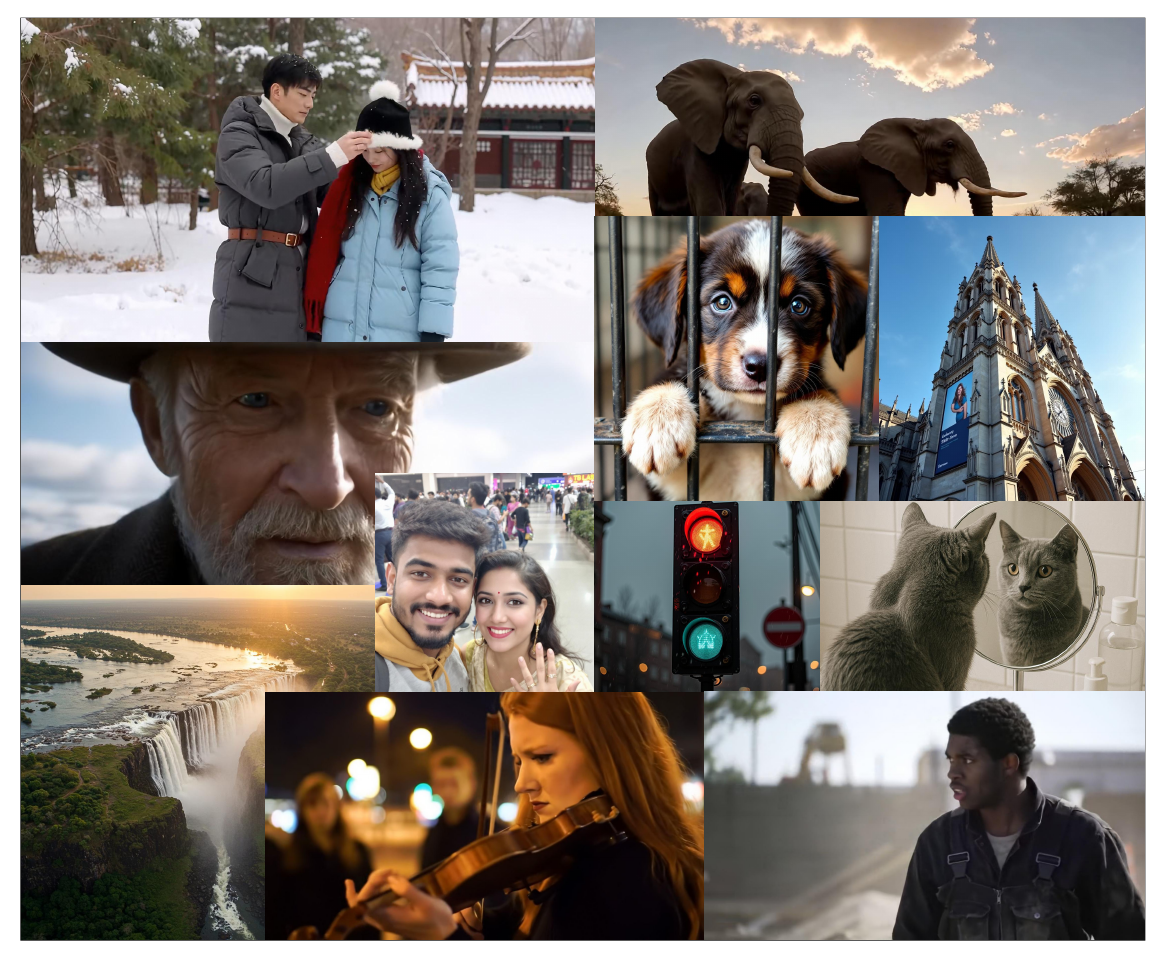}
    \caption{Visual Examples from GenBuster-Bench++.}
    \label{fig_6}
\end{figure}

\subsection{Motivation}
Studying whether image and video modalities exhibit constructive synergy or destructive interference calls for a unified evaluation platform with clearly specified sampling and screening criteria. Most existing AI-generated content benchmarks \cite{zhu2023genimagemillionscalebenchmarkdetecting, corvi2022detectionsyntheticimagesgenerated, dellanna2025truefakerealworldcase, xu2024fakeshield} remain confined to a single modality and fundamentally lack rigorous human curation, yielding superficial artifacts that are easily spotted by vanilla classifiers.

Naively merging these disjointed, single-modality datasets to evaluate unified MLLMs inevitably introduces profound experimental biases:
\textbf{1) Modality-Difficulty Imbalance:} If the synthetic video subset is elementary while the image subset is highly intricate, the model's joint optimization behavior will be heavily skewed, artificially masking the true cross-modal transfer mechanisms.
\textbf{2) Domain and Scenario Divergence:} Disjointed datasets feature distinct background distributions, semantic scenes, and context patterns, which forces the model to overfit to macroscopic content cues rather than learning underlying structural and pixel-level forgery invariants.
\textbf{3) Generator Generation Gaps:} Outdated generative patterns fail to capture the distribution of cutting-edge models, hindering investigations into how cross-modal knowledge generalizes to forgery patterns.

To support unified evaluation, we introduce \textbf{GenBuster-Bench++}. As illustrated in Figure \ref{fig:benchmark_overview}, the benchmark contains 4,000 curated samples, balanced across image and video modalities and real and synthetic classes. Both modalities cover the same scenario categories, and synthetic images and videos are screened using the same human difficulty criterion. This shared screening protocol provides a common inclusion criterion based on expert judgments for evaluating unified image-video detection.

\subsection{Benchmark Construction}

GenBuster-Bench++ leverages latest image and video generation techniques and consists of 4,000 images and video clips. It has two parts: real images/videos from real-world scenarios, and synthetic images/videos that simulate real-world conditions. The details are as follows:

\textbf{Data Sources:} We employ the following methods to construct a data pool for subsequent filtering.

{\it Real Images/Videos.} We source a large number of real images and videos from OpenVid-1M HD \cite{nan2024openvid}, a dataset that covers a diverse range of real-world scenarios. These samples are carefully pre-filtered to ensure they come from a wide variety of scenes.

{\it Fake Images/Videos.} We collect fake images and videos from multiple sources, including high-rated samples from MagicArena. In addition, we construct the following pipeline for generation: we use Reddit's official API to obtain social media images that cover a wide range of real-world scenarios. Then, we employ Qwen-2.5-VL \cite{bai2025qwen2} to generate detailed captions that describe the content of these source images. These captions then serve as prompts for various diffusion models to generate synthetic content. Finally, there are models such as FLUX \cite{labs2025flux1kontextflowmatching} and GPT-4o \cite{openai2024gpt4ocard} for image generation and Seedance 1.0 \cite{gao2025seedance10exploringboundaries}, SkyReels V1 \cite{SkyReelsV1} for video generation in the pool.

\textbf{Data Filtering:} At this stage, we introduce a rigorous filtering strategy to ensure that the final samples are of high-quality, diverse in content, and closely aligned with real-world scenarios.

\textit{Real Images/Videos.} We first filter out a large number of low-quality samples based on resolution, video frame rate, and video bitrate. Next, we eliminate duplicate content from the same origin clip. Finally, we manually check each sample one by one, removing those with extensive watermarks, anime content, or an obviously synthetic background. This process results in a final set of 1,000 real images and 1,000 real videos.

\textit{Fake Images/Videos.} To ensure the high quality and realism of fake samples, we implement a novel two-stage filtering pipeline. First, we create a mixed pool that contains both real and fake samples from the previous data pool. In the second stage, these identified samples are re-examined to separate those that were actually synthetic. We implemented a strict majority-vote protocol: a synthetic sample was retained in our final benchmark only when it successfully deceived at least 3 out of 5 experts into classifying it as real. This process results in a final set of 1,000 fake images and 1,000 fake videos.

\textbf{Post-processing.} For images, we standardize the resolution to 1024×1024. For video clips, we standardize the resolution to 1920×1080, the duration to 5 seconds, and the frame rate to 24 FPS. Figure \ref{fig_6} shows some samples. This unified post-processing approach offers several benefits: \textbf{1) Elimination of Encoding Biases}: By using HEVC encoding with x265, we eliminate potential biases from underlying different encoding preferences. \textbf{2) Consistency Across Sources}: Standardization ensures consistency across videos or images generated by different models, which may have varying original resolutions and frame rates.

\section{Method}
\label{sec:method}

In this section, we present \textbf{BusterX++}, a streamlined, single-stage RL trained MLLM for unified AI-generated image and video detection with autonomous explanation. Departing from traditional multi-stage pipelines, BusterX++ completely bypasses Supervised Fine-Tuning (SFT) to study joint image-video learning.

\subsection{Why Skip SFT before RL?}
Most existing reasoning-oriented MLLMs follow an
$\pi_{\text{base}}\rightarrow\pi_{\text{sft}}\rightarrow\pi_{\text{rl}}$
pipeline, where SFT initializes the policy using curated demonstrations before RL refinement.
However, applying this paradigm to AI-generated content detection introduces unique challenges. Reliable CoT supervision is difficult to obtain because explanations for visual forgery detection are often non-unique and depend on subtle visual cues such as textures, lighting, motion artifacts, and local inconsistencies. Human annotations may therefore exhibit substantial variance, while explanations synthesized by strong MLLMs can be post-hoc and weakly grounded in visual evidence.
As a result, SFT may bias optimization toward reproducing textual reasoning patterns instead of discovering discriminative visual evidence. Motivated by this observation, we investigate whether RL can be directly applied without SFT initialization, allowing the policy to explore task-specific reasoning behaviors through reward optimization.

\subsection{Unified Training}
We argue that combining image and video data can enhance unified learning through latent cross-modal complementarities. Specifically, high-quality static images provide rich, fine-grained texture and sharp boundary priors, which may conceptually assist the model in locating subtle, pixel-level forgery traces within video frames. Conversely, video sequences naturally encompass diverse compression noise and temporal variations, potentially enriching the model's physical robustness when evaluating subtle artifacts in single isolated images. By jointly training on both modalities within a shared optimization space, the model is expected to leverage such complementary information, thereby achieving stronger mutual generalization across both image and video forensics tasks. We adopt DAPO \cite{yu2025dapoopensourcellmreinforcement} as our RL strategy.

\textbf{Reward Functions.} The total reward is $R = r_{\text{format}} + r_{\text{accuracy}} + r_{\text{soft\_overlong}}$.

\begin{itemize}[leftmargin=*, nosep]
\item \textbf{Format}: If the model adheres to the specified format (i.e., \texttt{<think>...</think>\textbackslash n\textbackslash n...}), it receives a reward of $r_{\text{format}} = 0$; otherwise, it receives a penalty of $r_{\text{format}} = -1$.

\item \textbf{Accuracy}: If formatted and correct, it receives $r_{\text{accuracy}} = 1$; otherwise, $r_{\text{accuracy}} = 0$.

\item \textbf{Soft Overlong Penalty}: We apply a soft penalty based on the generated response length:

\begin{equation}
    r_{\text{soft\_overlong}} =
    \begin{cases}
    0, & L_{\text{gen}} \le L_{\text{eff}} \\
    \frac{L_{\text{eff}} - L_{\text{gen}}}{L_{\text{cache}}}, & L_{\text{eff}} <  L_{\text{gen}} \le L_{\text{max}} \\
    -1, & L_{\text{max}} < L_{\text{gen}}
    \end{cases}
\label{eq:soft_punish}
\end{equation}

\end{itemize}

\section{Experiments}
\label{sec:experiments}

\begin{figure*}[t]
    \centering
    \begin{subfigure}[b]{0.45\linewidth}
        \centering
        \includegraphics[width=\linewidth]{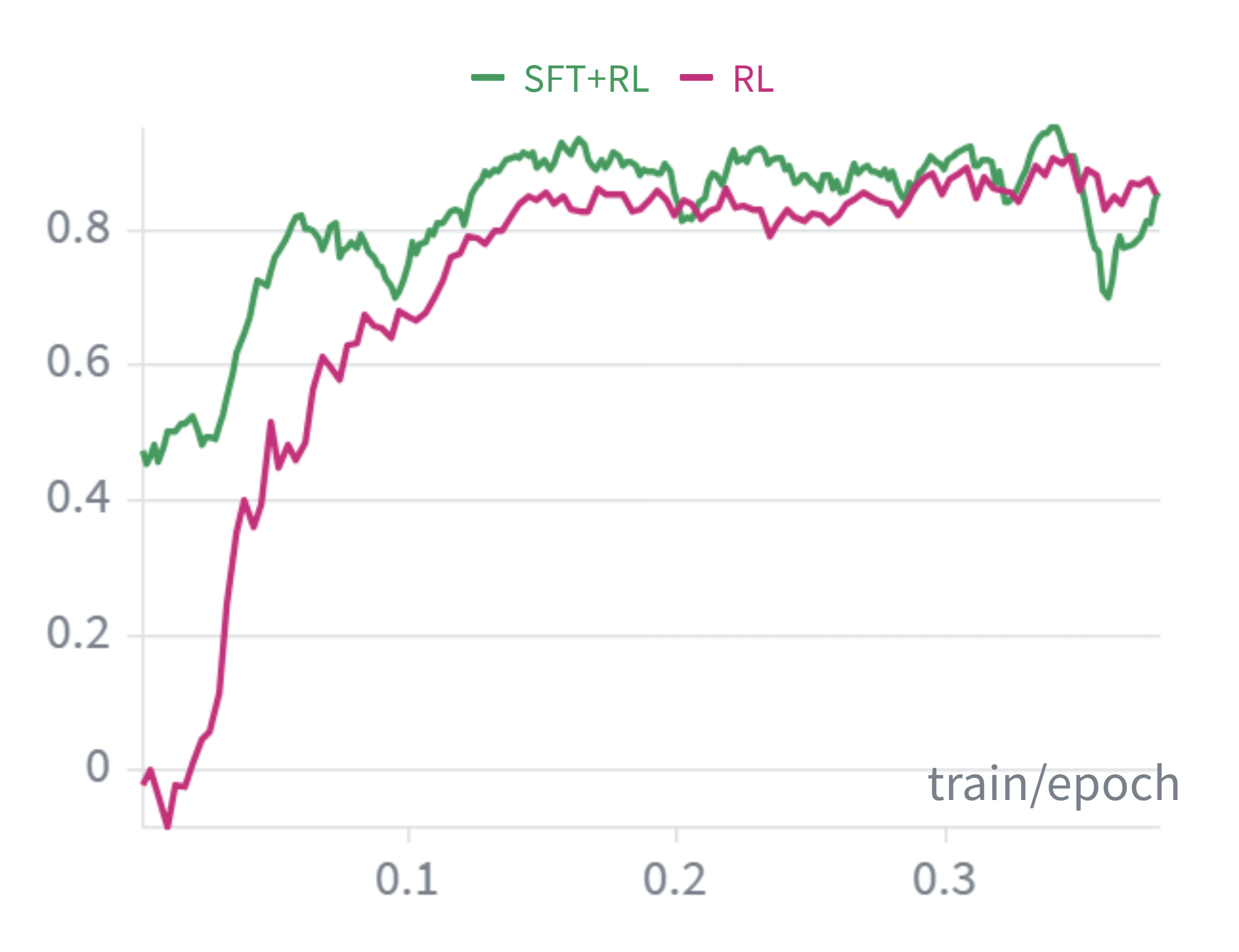}
        \vspace{-2em}
        \caption{Reward Dynamics}
        \label{fig:rl_reward}
    \end{subfigure}
    \hfill
    \begin{subfigure}[b]{0.45\linewidth}
        \centering
        \includegraphics[width=\linewidth]{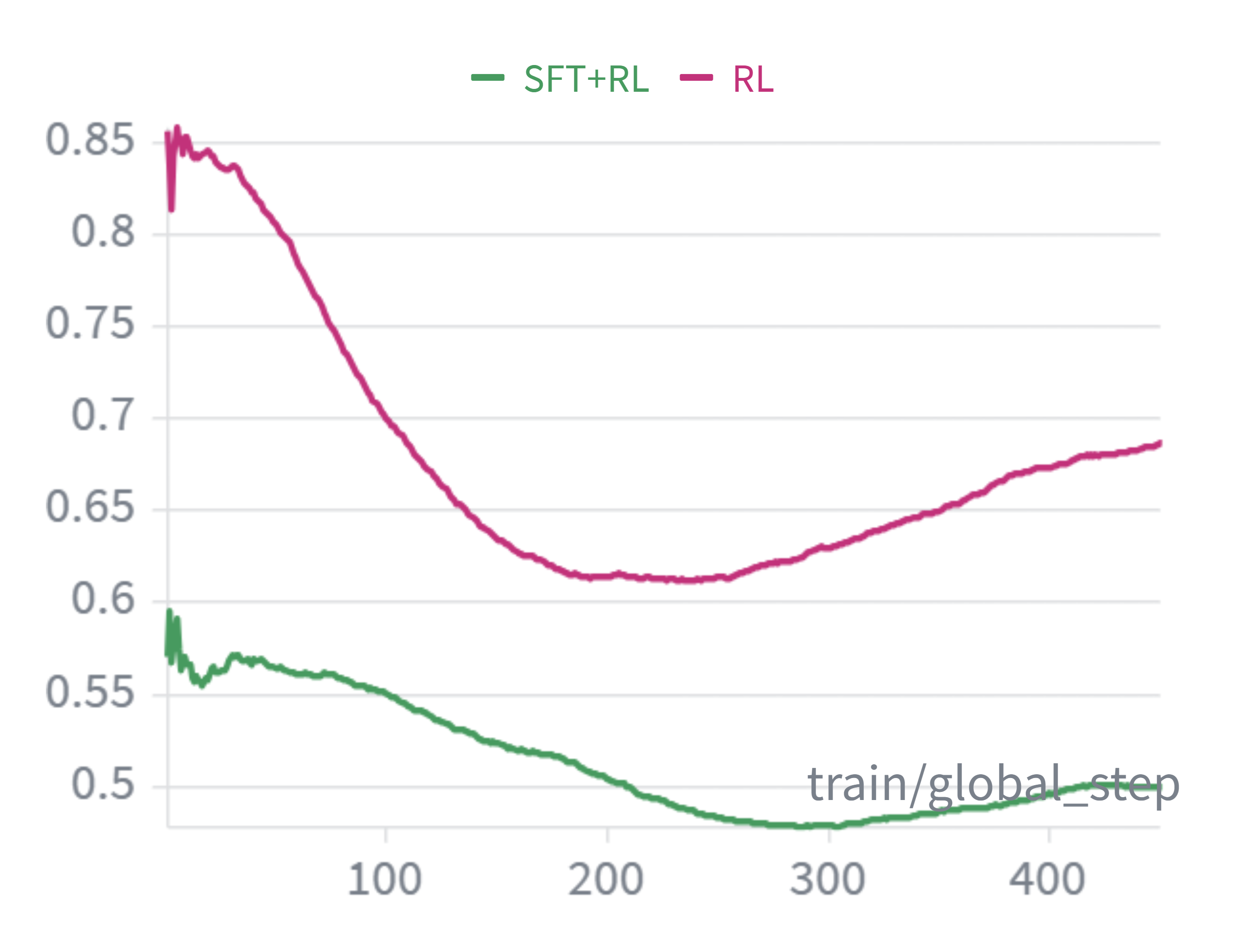}
        \vspace{-2em}
        \caption{Policy Entropy}
        \label{fig:rl_entropy}
    \end{subfigure}
    \vspace{-0.5em}
    \caption{Comparison of training pipelines between SFT-initialized RL (SFT+RL) and  single-stage RL. (a) Although the SFT baseline starts with a higher initial accuracy reward, the pure RL strategy aggressively optimizes from raw accuracy signals and matches the strong baseline as training progresses. (b) Pure RL maintains higher policy entropy than SFT-initialized RL throughout training.}
    \label{fig:pipeline_dynamics}
    \vspace{-0.5em}
\end{figure*}

\noindent
\textbf{Setup}.
We adopt Qwen3.5-4B \cite{qwen3.5} as our base model. For video-level forensic processing, we sample frames uniformly at a rate of 2.0 FPS. To optimize processing efficiency while preserving structural features, we enforce a strict token-level computational budget of 64 tokens per frame/image. 
We employ AdamW \citep{loshchilov2017decoupledadamw} with a learning rate of $1 \times 10^{-6}$, which aligns with established community best practices for stable RL training.
During training, we use DeepSpeed ZeRO and \texttt{bfloat16} precision. More details are shown in Appendix \ref{suppl:exp}.

\noindent
\textbf{Training Set}. 
To construct a computationally efficient yet modality-balanced training environment, we introduce \textbf{GenBuster-Unified}, a lightweight unified training set comprising 10,000 samples in total. The instances are sourced and curated from So-Fake-Set \cite{huang2025sofakebenchmarkingexplainingsocial} and GenBuster-200K \cite{wen2025busterxmllmpoweredaigeneratedvideo}. Notably, the generative models used to construct our evaluation benchmark are substantially \textbf{disjoint} from those in the training set, supporting evaluation of the framework's cross-generator generalization.

\begin{table}[htbp]
    \small
    \centering
    \caption{Comparison with existing MLLMs on GenBuster-Bench++.}
    \resizebox{0.48\textwidth}{!}{
    \begin{tabular}{l ccccc}
    \toprule
        \multirow{2}{*}{MLLMs} & \multicolumn{2}{c}{Image} & \multicolumn{2}{c}{Video} & \multirow{2}{*}{Overall} \\
        \cmidrule(r){2-3}
        \cmidrule(r){4-5}
        & Real & Fake & Real & Fake \\
        \midrule
        GLM-4.6V-Flash     & 96.8 & 02.3 & 91.0 & 32.4 & \cellcolor{lightgray}55.6 \\
        Qwen3.5-4B         & 25.6 & 11.5 & 85.6 & 27.7 & \cellcolor{lightgray}37.6 \\
        Qwen3.5-122B-A10B  & 82.1 & 23.9 & 83.4 & 32.8 & \cellcolor{lightgray}55.6 \\
        Qwen3.6-27B        & 80.0 & 27.7 & 87.3 & 34.8 & \cellcolor{lightgray}57.5 \\
        BusterX            & 82.1 & 68.3 & 88.7 & 73.4 & \cellcolor{lightgray}78.1 \\
        \midrule
        \textbf{BusterX++} & 85.4 & 76.3 & 90.5 & 74.8 & \cellcolor{lightgray}\textbf{81.8} \\
        \bottomrule
    \end{tabular}
    }
    \label{tab:genbuster_bench_plus_plus}
\end{table}

\begin{table}[htbp]
    \small
    \centering
    \caption{Ablation study on training pipelines and data modalities. Single-stage RL consistently outperforms SFT baselines, while cross-modal unified training improves over single-modality training under pure RL. All RL-stage runs use the same number of training steps.}
    \resizebox{0.48\textwidth}{!}{
    \begin{tabular}{ccccccccc}
    \toprule
        \multicolumn{2}{c}{\textbf{Training}} & \multicolumn{2}{c}{\textbf{Data}} & \multicolumn{2}{c}{\textbf{Image}} & \multicolumn{2}{c}{\textbf{Video}} & \multirow{2}{*}{\textbf{Overall}} \\
        \cmidrule(r){1-2} \cmidrule(r){3-4} \cmidrule(r){5-6} \cmidrule(r){7-8}
        SFT & RL & Image & Video & Real & Fake & Real & Fake & \\
        \midrule

        \ding{51} & -         & \ding{51} & - & 53.6 & 60.5 & 79.8 & 58.5 & \cellcolor{lightgray}63.1 \\
        \ding{51} & -         & - & \ding{51} & 48.0 & 56.1 & 91.0 & 66.7 & \cellcolor{lightgray}65.5 \\
        \ding{51} & -         & \ding{51} & \ding{51} & 64.4 & 59.7 & 90.8 & 63.5 & \cellcolor{lightgray}69.6 \\
        \ding{51} & \ding{51} & \ding{51} & \ding{51} & 84.2 & 63.6 & 93.4 & 70.1 & \cellcolor{lightgray}77.8 \\
        
        \midrule
        -         & \ding{51} & \ding{51} & - & 85.9 & 75.4 & 82.7 & 65.2 & \cellcolor{lightgray}77.3 \\
        -         & \ding{51} & -         & \ding{51} & 81.7 & 65.4 & 88.4 & 75.0 & \cellcolor{lightgray}77.6 \\
        -         & \ding{51} & \ding{51} & \ding{51} & 85.4 & 76.3 & 90.5 & 74.8 & \cellcolor{lightgray}\textbf{81.8} \\
    \bottomrule
    \end{tabular}
    }
    \label{tab:ablation}
\end{table}

\subsection{Performance on GenBuster-Bench++}
As shown in Table \ref{tab:genbuster_bench_plus_plus}, BusterX++ achieves notable overall accuracy. This indicates our model's strong generalization ability.

\begin{table*}[t]
    \centering
    \caption{Comparison of state-of-the-art methods on GenBuster-Bench. The benchmark comprises three progressive tracks: ID evaluates performance on seen generators; OOD assesses cross-generator generalization on unseen commercial models; and Wild stress-tests real-world robustness.}
    \small
    \resizebox{1\textwidth}{!}{
    \begin{tabular}{l ccc cccccccccc cc}
    \toprule
        \multirow{2}{*}{\textbf{Methods}} & \multicolumn{3}{c}{\textbf{ID (2024)}} & \multicolumn{10}{c}{\textbf{OOD (2025)}} & \multicolumn{2}{c}{\textbf{Wild (2026)}} \\
        \cmidrule(lr){2-4} \cmidrule(lr){5-14} \cmidrule(l){15-16}
        & Real & Fake & ACC & Sora & Pika & Gen3 & Luma & WanX & Kling & Jimeng & Vidu & Fake & ACC & Fake & ACC \\
        \midrule
        
        \rowcolor{lightgray} \multicolumn{16}{l}{\textit{\textbf{Classical Detectors}}} \\
        3D ResNeXt \cite{hara3dcnns}
        & 84.4 & 60.8 & 72.6 & 31.0 & 75.0 & 39.0 & 42.0 & 48.7 & 22.0 & 65.0 & 46.0 & 44.7 & 64.6 & 39.3 & 61.9 \\
        
        Vivit \cite{arnab2021vivit}
        & 92.9 & 64.1 & 78.5 & 74.0 & 52.0 & 69.0 & 53.0 & 62.0 & 55.0 & 33.0 & 68.7 & 60.6 & 76.8 & 46.7 & 69.8 \\
        
        VideoMAE \cite{tong2022videomae}
        & 93.3 & 64.9 & 79.1 & 48.5 & 79.0 & 65.0 & 52.0 & 76.7 & 24.0 & 55.0 & 72.0 & 59.5 & 76.4 & 50.7 & 72.0 \\
        
        DeMamba \cite{chen2024demamba}
        & \textbf{93.8} & 70.2 & 82.0 & 65.0 & 75.0 & 73.0 & 56.0 & 46.0 & 68.0 & 71.0 & 84.7 & 66.9 & 80.4 & 48.0 & 70.9 \\
        
        \midrule
        \rowcolor{lightgray} \multicolumn{16}{l}{\textit{\textbf{Closed-Source MLLMs}}} \\
        GPT-5.2 (High) \cite{gpt5.2}
        & 93.3 & 61.8 & 77.6 & 43.0 & 77.0 & 78.0 & 91.0 & 66.7 & 36.0 & 27.0 & 33.3 & 54.5 & 73.9 & 50.7 & 72.0 \\
        
        Grok-4.1-fast \cite{grok4.1_fast}
        & 76.1 & 77.5 & 76.8 & 78.0 & 81.0 & 60.0 & 90.0 & 80.0 & 87.0 & 92.0 & 73.3 & 79.6 & 77.9 & 35.3 & 55.7 \\
        
        Claude-Haiku-4.5 \cite{claude-haiku-4-5}
        & 84.7 & 58.7 & 71.7 & 53.0 & 40.0 & 72.0 & 80.0 & 60.0 & 31.0 & 46.0 & 46.7 & 53.5 & 69.1 & 19.3 & 52.0 \\
        
        Claude-Sonnet-4.6 \cite{claude-sonnet-4-6}
        & 61.6 & 83.2 & 72.4 & 84.0 & 91.0 & 93.0 & 94.0 & 93.3 & 72.0 & 61.0 & 80.0 & 83.9 & 72.8 & 63.3 & 62.5 \\

        \midrule 
        \rowcolor{lightgray} \multicolumn{16}{l}{\textit{\textbf{Open-Source MLLMs}}} \\
        Qwen3.5-4B \cite{qwen3.5}
        & 83.1 & 48.9 & 66.0 & 30.0 & 69.0 & 68.0 & 70.0 & 52.7 & 44.0 & 43.0 & 60.7 & 52.4 & 67.8 & 29.3 & 56.2 \\
        
        Qwen3.5-27B \cite{qwen3.5}
        & 76.3 & 54.8 & 65.6 & 43.0 & 70.0 & 72.0 & 89.0 & 61.3 & 47.0 & 43.0 & 54.0 & 58.0 & 67.2 & 38.0 & 57.2 \\
        
        Qwen3.5-122B-A10B \cite{qwen3.5}
        & 79.4 & 54.7 & 67.1 & 57.0 & 87.0 & 73.0 & 93.0 & 66.0 & 49.0 & 40.0 & 68.0 & 65.7 & 72.6 & 42.0 & 60.7 \\
        
        \midrule
        \rowcolor{lightgray} \multicolumn{16}{l}{\textit{\textbf{domain-specific MLLMs}}} \\

        \textbf{BusterX}
        & 83.0 & 90.4 & 86.7 & 91.0 & 89.0 & 92.0 & 99.0 & 94.0 & 93.0 & 87.0 & 94.7 & 92.5 & 87.8 & \textbf{72.0} & 77.5 \\
        
        \midrule
        \textbf{BusterX++}
        & 87.0 & \textbf{91.6} & \textbf{89.3} & \textbf{96.0} & \textbf{100.0} & \textbf{94.0} & \textbf{100.0} & \textbf{98.0} & \textbf{98.0} & \textbf{96.0} & \textbf{98.0} & \textbf{97.4} & \textbf{92.2} & 70.7 & \textbf{78.9} \\
        \bottomrule
    \end{tabular}
    }

    \raggedright
    \scriptsize{\textsuperscript{\dag} \textit{Note:} The ACC for the OOD and Wild tracks is calculated as the macro-average of the track's fake accuracy and the ID Real accuracy.}
    \label{tab:genbuster_bench}
\end{table*}

\subsection{Performance on GenBuster-Bench}
We further evaluate the generalization capabilities of BusterX++ on GenBuster-Bench \cite{wen2025busterxmllmpoweredaigeneratedvideo}, as comprehensively detailed in Table~\ref{tab:genbuster_bench}. BusterX++ is competitive with both closed-source and open-source models, delivering exceptional performance under out-of-distribution and wild setups.

\subsection{Comparison with Strong SFT Baseline}
To benchmark the conventional \emph{SFT $\rightarrow$ RL} pipeline, we construct a strong SFT baseline backed by a rigorous data synthesis pipeline. Specifically, we harvest extra 10,000 high-quality, long-form reasoning rationales (CoT) distilled from Qwen3.5-397B in So-Fake-Set \cite{huang2025sofakebenchmarkingexplainingsocial} and GenBuster-200K \cite{wen2025busterxmllmpoweredaigeneratedvideo}. To ensure the synthetic data possesses both rigorous forensic depth and precise structural alignment, we feed these images and video frames into the teacher model along with a carefully engineered prompt. This prompt explicitly instructs the teacher model to execute a three-stage pipeline: (1) systematically isolate low-level perceptual anomalies, (2) cross-verify these findings against semantic world knowledge, and (3) output a structured, step-by-step reasoning chain. To filter out potential teacher hallucinations, we implement an automated validation pass, retaining only those synthetic samples where the teacher's final boxed answer aligns perfectly with the absolute ground-truth labels. The base model is then fine-tuned on this 10,000-sample corpus via standard SFT for 1 epoch.

We then track the RL training dynamics of both configurations in Figure~\ref{fig:pipeline_dynamics}. Although the SFT baseline inherently starts with a higher accuracy reward, our single-stage RL approach exhibits a remarkably steep optimization curve. Driven strictly by sparse accuracy rewards without any SFT warm-up, the pure RL policy steadily catches up with and eventually outshines the strong SFT+RL pipeline in the later phase. This phenomenon is explicitly explained by the policy entropy dynamics in Figure~\ref{fig:rl_entropy}. Strong SFT constrains the policy search space and dampens exploratory freedom. In stark contrast, our single-stage pure RL paradigm sustains significantly higher policy entropy throughout training. By keeping the reasoning paths unconstrained, the model is compelled to actively explore and locate implicit visual anomalies embedded within pixel distributions.

Table~\ref{tab:ablation} further compares the detection accuracy of the two pipelines. Under the joint image-video training mode, our pure RL strategy achieves accuracy of \textbf{81.8\%}, significantly outperforming the standard SFT+RL joint training baseline (77.8\%).

\begin{figure*}[t]
    \centering
    \includegraphics[width=1\textwidth]{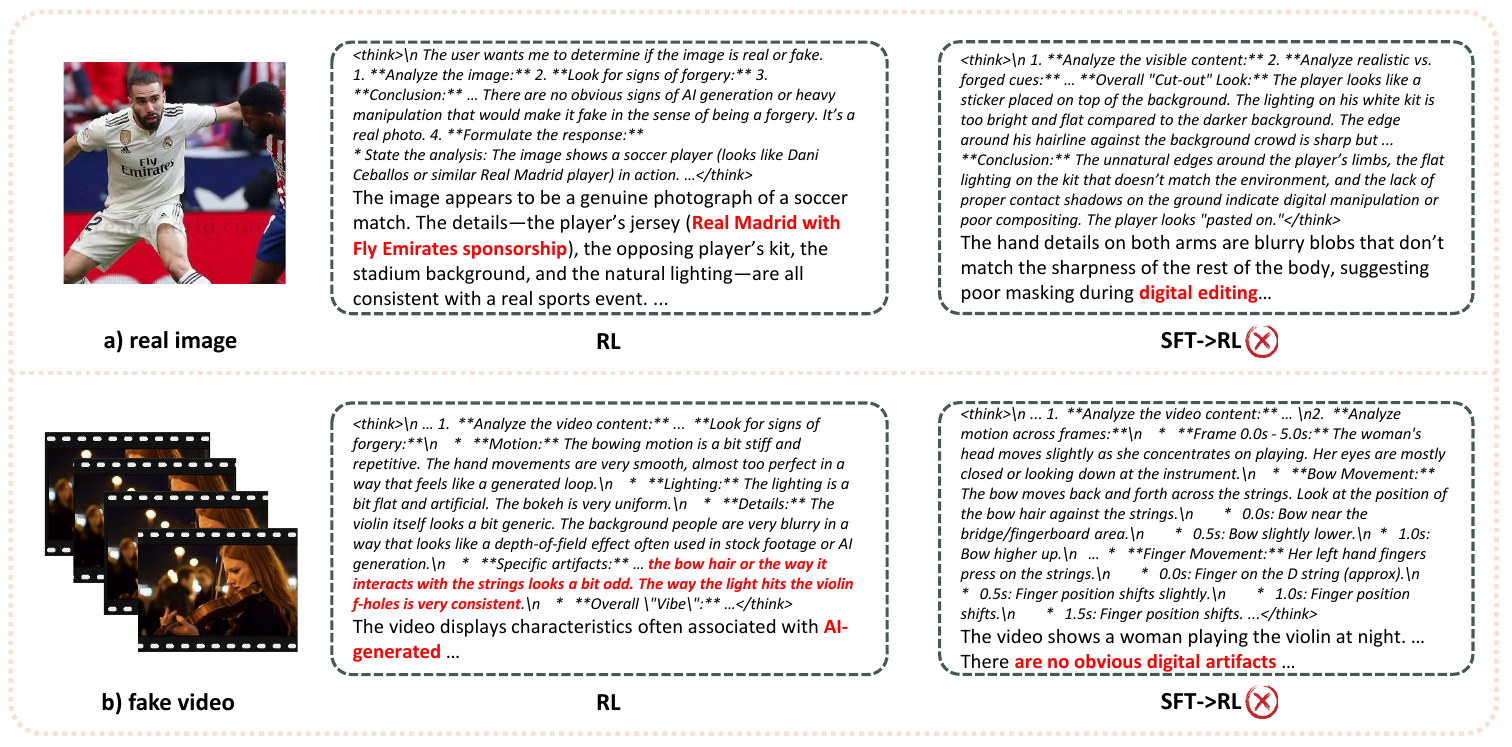}
    \caption{Two special cross-modal transfer cases between the SFT$\rightarrow$RL baseline and our single-stage pure RL paradigm. (a) The SFT baseline mistakes dynamic motion blur for digital compositing, whereas pure RL correctly identifies it as a real sports photograph. (b) The SFT baseline is deceived by globally consistent lighting, while pure RL exposes the forgery by capturing frame-local appearance artifacts. Full responses are provided in Appendix \ref{suppl:complete}.}
    \label{fig:cross_modal_transfer}
\end{figure*}

\subsection{Analysis of Unified Training}
To investigate our hypothesis that unified training improves cross-modal learning, we compare the unified model against single-modality models. As shown in Table~\ref{tab:ablation}, models trained on a single modality underperform the unified approach even within their own native modality, highlighting the strict limitations of isolated training. Crucially, whether deploying SFT or RL, joint image-video training consistently yields significant numerical improvements in overall accuracy. This empirical gain confirms that joint training successfully leverages complementary cross-modal information, creating a virtuous cycle that enhances global generalization.

To further examine the possible mechanics of this cross-modal capability transfer, we perform a fine-grained case study comparing the SFT+RL unified model against pure RL in Figure~\ref{fig:cross_modal_transfer}. These cases illustrate differences in the visual cues cited by the two models:

\begin{enumerate}

\item \textbf{Video-to-Image Robustness Transfer}: Conversely, extensive training on dynamic video sequences equips the pure RL model with an intrinsic tolerance to temporal ambiguities, which directly benefits static image detection—especially for fast-action video screenshots. As shown in Figure~\ref{fig:cross_modal_transfer}a, when evaluating a real soccer match photograph, the SFT baseline overreacts to the intense motion blur and stretched limbs, misinterpreting natural photographic blur as poor masking and a "cut-out" digital edit. Conversely, pure RL seamlessly applies its video-learned dynamic-scene priors to the single image. It recognizes that localized blur and edge softness are characteristic of high-speed real-world sports capture, thereby maintaining robustness and correctly classifying the frame as authentic.

\item \textbf{Image-to-Video Feature Transfer}: While the conventional SFT+RL baseline excels at learning globally coherent physical narratives, it often over-trusts macro-level consistency. As illustrated in Figure~\ref{fig:cross_modal_transfer}b, while the SFT baseline over-relies on global temporal dynamics and misclassifies the sample, our pure RL strategy prioritizes frame-local artifacts. This image-level sensitivity transfers to the video domain, enabling the model to bypass global coherence and detect the forgery through localized anomalies.

\end{enumerate}

\begin{table}[htbp]
    \small
    \centering
    \caption{Robustness evaluation of BusterX++.}
    \resizebox{0.30\textwidth}{!}{
    \begin{tabular}{cccc}
    \toprule
        \multicolumn{3}{c}{Perturbation} & \multirow{2}{*}{Overall} \\
        \cmidrule(r){1-3}
        JPEG & Noise & Blur & \\
        \midrule
        -         & -         & -         & 81.8 \\
        \ding{51} & -         & -         & 78.9 \\
        -         & \ding{51} & -         & 76.1 \\
        -         & -         & \ding{51} & 80.4 \\
        \ding{51} & \ding{51} & \ding{51} & 75.3 \\
    \bottomrule
    \end{tabular}
    }
    \label{tab:robustness}
\end{table}

\subsection{Robustness Study}
We further assessed the robustness of BusterX++ against common visual perturbations to simulate real-world scenarios, including JPEG Compression (quality=70), Gaussian Noise ($\sigma$=5), Gaussian Blur, and their Degradation Cascade \cite{wang2021realesrgan}.
As shown in Table~\ref{tab:robustness}, BusterX++ shows remarkable stability against these low-level distortions, even without specific training on degraded data.
Gaussian Blur appears to have only a marginal effect on overall accuracy. BusterX++ still delivers solid performance under the degradation cascade, highlighting its robustness and practical value in diverse and unpredictable real-world settings.

\subsection{Case Study}
We provide a qualitative review to showcase the model's performance in distinguishing between real and fake samples. Due to space constraints, additional qualitative examples and case studies are presented in Appendix~\ref{suppl:complete}, further demonstrating the effectiveness of BusterX++ across diverse scenarios.

\begin{enumerate}

\item \textbf{Stable Reasoning}: Across all evaluated scenarios, BusterX++ consistently adopts a structured, step-by-step analytical pattern instead of producing a superficial, binary judgment. As detailed in Figure~\ref{fig:cross_modal_transfer}, the model systematically structures its \texttt{<think>} trajectory. It explicitly enumerates localized elements, showing exceptional reasoning stability regardless of the task modality.

\item \textbf{Fine-grained Visual Perception}: The pure RL paradigm grants the model a heightened sensitivity toward micro-level, low-level details that are easily overlooked by traditional SFT-initialized pipelines. For instance, in the image-to-video transfer scenario (Figure~\ref{fig:cross_modal_transfer}b), BusterX++ bypasses the deceptive global aesthetic to expose the forgery, showing robust perception of fine-grained artifact invariants.

\item \textbf{World Knowledge}: Beyond tracking statistical pixel-level footprint anomalies, BusterX++ demonstrates an elegant capacity for leveraging internal parametric world knowledge to cross-verify physical and semantic plausibility. In Figure~\ref{fig:cross_modal_transfer}a), it utilizes deep context familiarity to verify the authentic coexistence of official Real Madrid kit markers alongside correct sponsor typography.

\end{enumerate}

\subsection{User Preference Evaluation}
Evaluating explanations for visual AIGC detection is challenging due to the lack of canonical ground-truth rationales. To address this, we conduct a strict double-blind user preference evaluation to compare the explanation quality of our model with GPT-5.2. The evaluation set contains 100 AI-generated samples, including 50 images and 50 videos, which are mixed during the rating phase. 
The evaluation was conducted by the exact same panel of 5 forensic experts who filtered our benchmark. For each sample, experts were blindly presented with the media and two anonymized candidate explanations (Ours vs. GPT-5.2) in a randomized order. Each expert cast an independent vote evaluating multiple analytical dimensions, including evidence grounding, specificity, and completeness, with the winning explanation determined by a strict 3-out-of-5 majority consensus. BusterX++ explanations received a majority preference over GPT-5.2 explanations on \textbf{82\%} of these 100 samples.

Furthermore, we conducted a factual grounding check to assess whether the visual artifacts cited in the model's intermediate explanations were visible in the media. The experts blindly audited 100 correctly identified fake cases. Under the same majority-vote protocol, they judged the cited artifacts in BusterX++'s \texttt{<think>} blocks (e.g., texture blending and reflection anomalies) to be present in \textbf{87\%} of these cases.

\section{Conclusion}
\label{sec:conclusion}

In this paper, we present \textbf{BusterX++}, a unified MLLM that utilizes a single-stage RL strategy optimized directly via sparse accuracy rewards for cross-modal AI-generated content detection. To facilitate evaluation, we introduce \textbf{GenBuster-Bench++}, a curated unified benchmark consisting of 4,000 image and video samples. We observe higher policy entropy under direct RL during post-training, alongside improved overall detection accuracy under joint image-video training. Comprehensive evaluations demonstrate that our approach yields superior or competitive performance compared to strong SFT-initialized baselines, highlighting the potential of RL for cross-modal reasoning.

\noindent\textbf{Societal Impact.} AI-generated content detection models are not infallible: errors in classification or explanation could lead to false positives, incorrectly labeling authentic content as synthetic, or false negatives, failing to flag harmful AI-generated content. We encourage the adoption of these models alongside human oversight, transparent reporting, and continuous evaluation to ensure that the benefits of AI-generated content detection outweigh potential harms. By documenting both capabilities and limitations, our work aims to provide a balanced perspective on the societal impact of reasoning-driven, multimodal AI-generated content detection.

% \newpage
\section*{Limitations}
\label{sec:limitations}

Despite promising progress, we recognize certain limitations and aim to address them through future optimization. \textbf{1) Generative Technology Adaptation.} We observe that samples generated with the latest technologies in GenBuster-Bench++ present a more significant challenge. This indicates that the rapid evolution of generative methods necessitates continuous adaptation and enhancement of our detection framework to maintain its efficacy. \textbf{2) Potential Bottleneck in Post-training.} Post-training phase may be approaching a performance bottleneck in our task. To further advance the capabilities of MLLMs, future work should explore other training stages.

\section*{Acknowledgments}
This work was funded by the Engineering and Physical Sciences Research Council [grant number EP/X525741/1] Impact Acceleration Accounts Grant.

% \newpage
\bibliography{main}

% \newpage
\appendix
\section{DAPO}
\label{suppl:dapo}

We adopt DAPO \cite{yu2025dapoopensourcellmreinforcement} as our RL strategy. DAPO samples a group of outputs $\{o_i\}_{i=1}^G$ for each question $q$ from the old policy $\pi_{\theta_{old}}$. The reward model is then used to score the outputs, yielding $\{r_i\}_{i=1}^G$ correspondingly. Then DAPO optimizes the policy model $\pi_{\theta}$ by maximizing the following objective:

\begin{equation}
\begin{aligned}
&\mathcal{J}_{\text{DAPO}}(\theta) = \mathbb{E}_{(q,a)\sim \mathcal{D}, \{o_i\}_{i=1}^G\sim \pi_{\theta_\text{old}}(\cdot\mid q)}\\&
\Bigg[\frac{1}{\sum_{i=1}^{G}|o_i|}\sum_{i=1}^{G}\sum_{t=1}^{|o_i|} 
\min \Big( r_{i,t}(\theta) \hat{A}_{i,t},\\&
\text{clip} \Big( r_{i,t}(\theta), 1 - {\varepsilon_{\text{low}}}, 1 + {\varepsilon_{\text{high}}} \Big) \hat{A}_{i,t} \Big) \Bigg]
\\
\text{s.t.}\quad& 0< \Big|\{o_i\mid\text{is\_equivalent}(a,o_i)\}\Big|< G,
\label{eq:dapoloss}
\end{aligned}
\end{equation}
where $\varepsilon_{\text{low}}$ and $\varepsilon_{\text{high}}$ are hyper-parameters, and $\hat{A}_{i,t}$ is the advantage:

% \begin{equation}
%     r_{i,t}(\theta)=\frac{\pi_{\theta}(o_{i,t} \mid q, o_{i,<t})}{\pi_{\theta_{\text{old}}}(o_{i,t} \mid q,o_{i,<t})},\quad\hat{A}_{i,t} = \frac{r_i - \text{mean}(\{r_i\}_{i=1}^G)}{\text{std}(\{r_i\}_{i=1}^G)}
% \label{eq:advantage_calculation}
% \end{equation}

\begin{equation}
    r_{i,t}(\theta)=\frac{\pi_{\theta}(o_{i,t} \mid q, o_{i,<t})}{\pi_{\theta_{\text{old}}}(o_{i,t} \mid q,o_{i,<t})}
\label{eq:advantage_calculation_1}
\end{equation}

\begin{equation}
    \quad\hat{A}_{i,t} = \frac{r_i - \text{mean}(\{r_i\}_{i=1}^G)}{\text{std}(\{r_i\}_{i=1}^G)}
\label{eq:advantage_calculation_2}
\end{equation}

\section{Experimental Details}
\label{suppl:exp}

In this section, we provide comprehensive details regarding the training configurations, evaluation parameters, and hyperparameter selections utilized in our study to ensure full reproducibility.

To minimize generation overhead, we integrate \texttt{vLLM} \citep{kwon2023vllm} for both training rollouts and evaluation. Its high-throughput capabilities enable us to maintain a large training rollout batch size. For each query, we generate $G=8$ trajectories at a temperature of $T=1.0$. The maximum completion length is set to $4,096$ tokens. We apply a clipping bound of $[0.8, 1.28]$ for the policy ratio.

\section{Prompt Design}
\label{suppl:prompt}

Based on extensive experiments, we have carefully designed the following prompt to guide the answer.

\begin{tcolorbox}[title = {User Prompt -- Video}, fontupper=\small]
{
\it 
Please analyze whether there are any inconsistencies or obvious signs of forgery in the video, and finally come to a conclusion: Is this video real or fake?\\

Then, just answer this MCQ with a single letter:\\
Q: Is this video real or fake?\\
Options:\\
A) real\\
B) fake\\

Put the final answer in $\backslash \backslash boxed\{...\}$
}
\end{tcolorbox}

\begin{tcolorbox}[title = {User Prompt -- Image}, fontupper=\small]
{
\it 
Please analyze whether there are any inconsistencies or obvious signs of forgery in the image, and finally come to a conclusion: Is this image real or fake?\\

Then, just answer this MCQ with a single letter:\\
Q: Is this image real or fake?\\
Options:\\
A) real\\
B) fake\\

Put the final answer in $\backslash \backslash boxed\{...\}$
}
\end{tcolorbox}

\section{Details of User Preference Evaluation}
\label{suppl:pref_eval}
We conduct a strict double-blind user preference evaluation to compare the explanation quality of our model with GPT-5.2. For each sample, experts are instructed to judge explanations from these perspectives: evidence grounding, meaning claims are supported by visible cues; specificity, meaning clear and concrete cues rather than vague statements; internal consistency, meaning no contradictions across points; completeness and coverage, meaning it covers major relevant visual and temporal aspects; and clarity and usability, meaning it is easy to follow and helps a reader verify cues.

% pppppppppppppppppppppppppppppppppppppppppppppppppppppppp

\onecolumn
\newpage

\section{Complete Responses}
\label{suppl:complete}

\begin{figure}[htbp]
  \centering
  \includegraphics[width=0.3\linewidth]{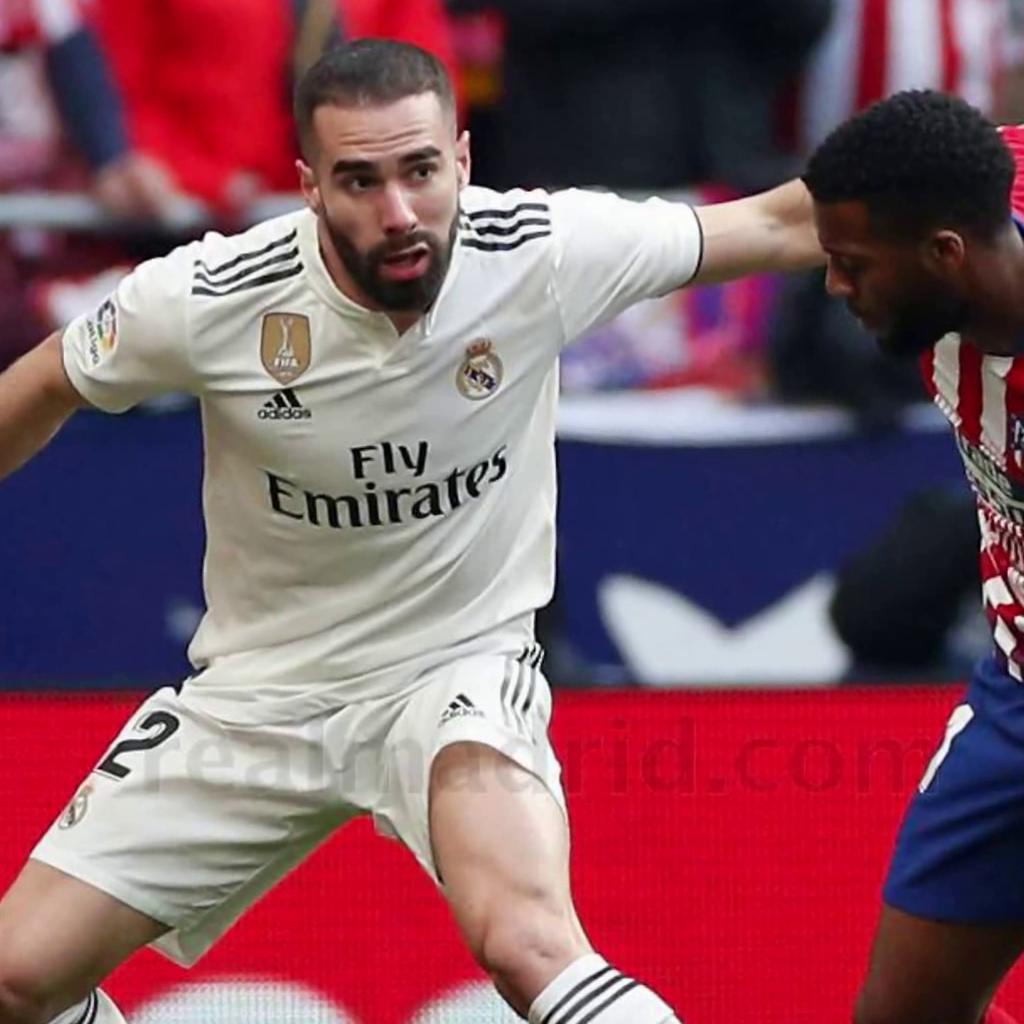}
\end{figure}

\begin{tcolorbox}[title = {Analysis of Unified Training -- REAL IMAGE (SFT+RL) }, fontupper=\tiny]
{
<think>\\The user wants me to determine if the image is real or fake.\\\\1.  **Analyze the visible content:**\\    *   **Subject:** A soccer player in a white kit (Real Madrid jersey) with Fly Emirates sponsorship. He has a beard and short hair.\\    *   **Action:** He appears to be in motion, possibly dribbling or shielding the ball, with his left arm extended.\\    *   **Opponent:** Another player in a blue and red striped kit (Atletico Madrid) is partially visible on the right edge.\\    *   **Background:** Blurry crowd and stadium seating, typical of a broadcast feed.\\\\2.  **Analyze realistic vs. forged cues:**\\    *   **Lighting and Shadows:** Look at the light source. It seems to be coming from above (stadium lights). Shadows on the ground should be consistent. Look at the shadow under the player's right foot (bottom left). It looks like a dark blob on the red turf, but it's very indistinct and doesn't match the light direction perfectly.\\    *   **Texture of the Kit:** Look closely at the fabric of the white jersey. It looks overly smooth and plastic-like in some areas, lacking the natural creasing and weave of cotton/polyester fabric usually seen in high-res sports photography. The folds around the waistband look painted on rather than folded.\\    *   **Edges and Boundaries:** Look where the player meets the background.\\        *   **Left Arm:** The hand on the left side (viewer's left) looks strange. The fingers are elongated and blurry in a way that suggests digital smearing or poor masking during cut-and-paste editing. The wrist angle looks awkward against the shoulder.\\        *   **Right Arm:** The hand on the right side (viewer's right) is extended towards the opponent. Look at the fingers. They look like a blurry blob. The connection to the elbow looks stiff.\\        *   **Legs:** Look at the right leg (viewer's left). The knee joint looks weirdly bent. The sock/shin guard area blends into the shorts without clear definition. Look at the left leg (viewer's right). The foot is planted, but the shadow underneath it is just a dark line.\\    *   **Lighting on the Face:** The light on the player's face (forehead, nose) seems flat. It doesn't match the strong light usually seen on a stadium pitch. The shadow under the chin looks disconnected from the neck.\\    *   **Inconsistencies with the Opponent:** Look at the opponent's arm. It extends into the frame. The hand looks like a dark blob. The connection point where the opponent's arm meets the shoulder looks like a dark smear.\\    *   **Overall "Cut-out" Look:** The player looks like a sticker placed on top of the background. The lighting on his white kit is too bright and flat compared to the darker background. The edge around his hairline against the background crowd is sharp but lacks the natural halo or blur that would occur from depth of field. The edge around his left arm (viewer's left) against the background is also sharp and unnatural. It looks like a composite image where a player was cut out and pasted onto a background plate without proper lighting integration. The shadow on the ground (bottom left) doesn't anchor him to the field; he looks like he's floating slightly above it.\\\\3.  **Conclusion:** The unnatural edges around the player's limbs, the flat lighting on the kit that doesn't match the environment, and the lack of proper contact shadows on the ground indicate digital manipulation or poor compositing. The player looks "pasted on."\\</think>\\\\The image shows a soccer player in a white Real Madrid kit, but several cues indicate manipulation. The lighting on the player's white jersey is flat and bright, contrasting with the darker, shadowed background of the stadium stands. More critically, look at the edges where the player meets the background: his left arm (on the left side of the image) and right arm (on the right side) have sharp, unnatural boundaries that look like a cutout. He appears to be floating above the red turf rather than standing on it, as there is no clear contact shadow beneath his feet to ground him in the scene. The hand details on both arms are blurry blobs that don't match the sharpness of the rest of the body, suggesting poor masking during digital editing.\\

$\backslash \backslash boxed\{B\}$
}
\end{tcolorbox}

\newpage
\begin{figure}[htbp]
  \centering
  \includegraphics[width=0.3\linewidth]{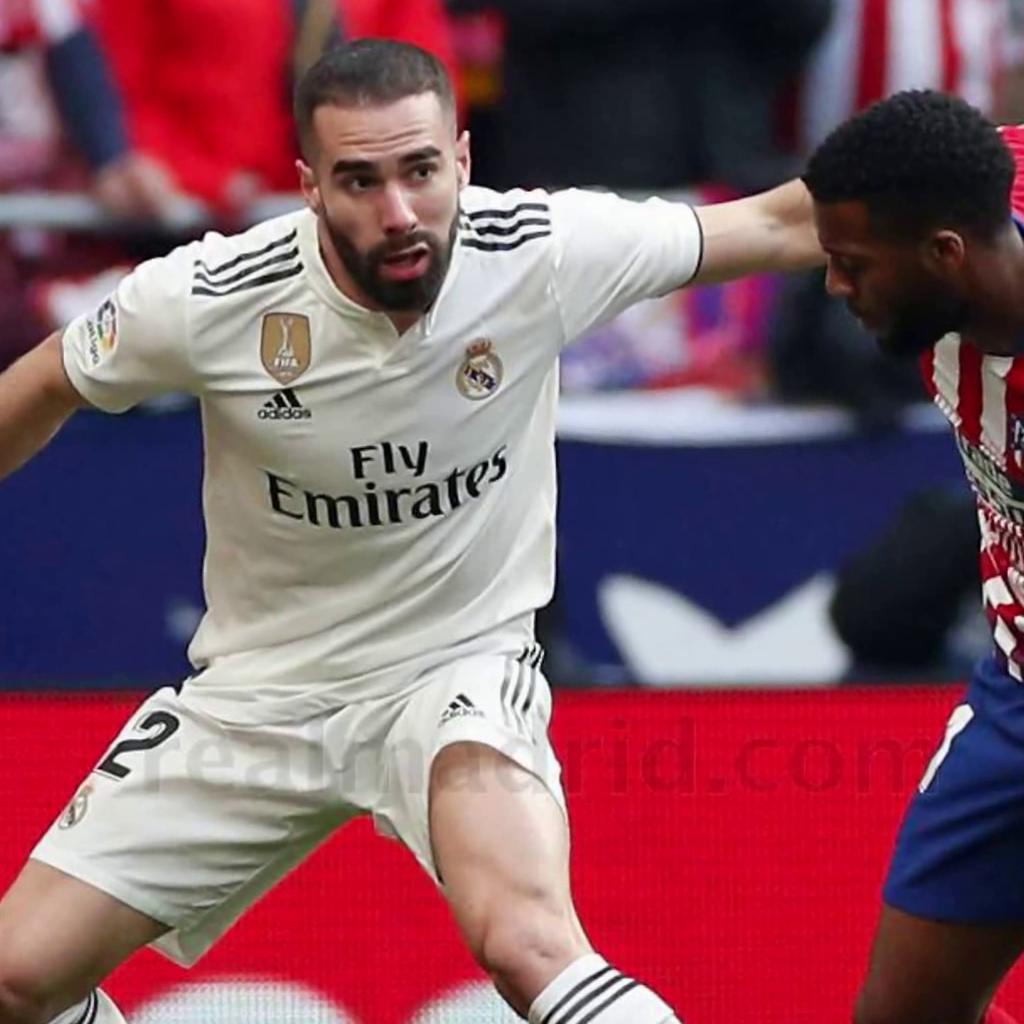}
\end{figure}

\begin{tcolorbox}[title = {Analysis of Unified Training -- REAL IMAGE (RL) }, fontupper=\tiny]
{
<think>\\The user wants me to determine if the image is real or fake.\\\\1.  **Analyze the image:**\\    *   **Subject:** A soccer player in a white kit (Real Madrid, Fly Emirates sponsor, Adidas logo). He has a beard and short hair.\\    *   **Action:** He is running or dribbling, looking focused.\\    *   **Background:** Another player in a red and white striped kit (Atletico Madrid, likely) is partially visible on the right. There's a red advertising board in the background.\\    *   **Details:**\\        *   The jersey has the Real Madrid crest.\\        *   The sponsor Fly Emirates is clear.\\        *   The Adidas stripes are visible.\\        *   There's a watermark id.com visible across the bottom.\\        *   The lighting looks like stadium lighting.\\        *   The textures (fabric, skin) look natural.\\\\2.  **Look for signs of forgery:**\\    *   **Distortion:** Are limbs twisted unnaturally? No, the pose looks like a legitimate athletic pose.\\    *   **Artifacts:** Are there weird blending edges? No.\\    *   **Text:** Is the text garbled? Fly Emirates is clear. The watermark is a standard stock photo watermark.\\    *   **Consistency:** The colors are consistent with a real sports broadcast.\\\\3.  **Conclusion:** The image looks like a genuine photograph from a soccer match. It's a standard sports action shot. There are no obvious signs of AI generation or heavy manipulation that would make it fake in the sense of being a forgery. It's a real photo.\\\\4.  **Formulate the response:**\\    *   State the analysis: The image shows a soccer player (looks like Dani Ceballos or similar Real Madrid player) in action. The details like the jersey, logos, and background are consistent with a real sports photograph.\\    *   Answer the MCQ: A) real.\\</think>\\\\The image appears to be a genuine photograph of a soccer match. The details—the player's jersey (Real Madrid with Fly Emirates sponsorship), the opposing player's kit, the stadium background, and the natural lighting—are all consistent with a real sports event. There are no obvious artifacts or distortions typical of image forgery.\\

$\backslash \backslash boxed\{A\}$
}
\end{tcolorbox}

\newpage
% image here
\begin{figure}[htbp]
  \centering
  \includegraphics[width=1\linewidth]{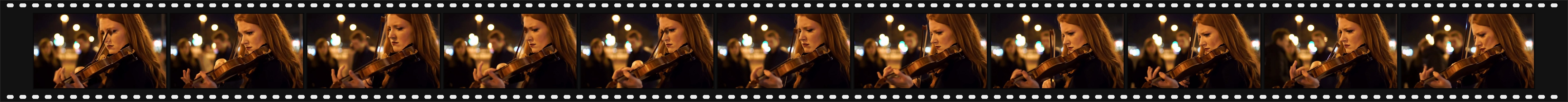}
\end{figure}

\begin{tcolorbox}[title = {Analysis of Unified Training -- FAKE VIDEO (SFT+RL)}, fontupper=\tiny]
{
<think>\\The user wants me to analyze a video and determine if it's real or fake.\\\\1.  **Analyze the video content:**\\    *   **Subject:** A woman with long reddish-brown hair playing a violin. She is in the foreground, right side of the frame.\\    *   **Action:** She is bowing the violin. Her left hand is on the fingerboard, pressing strings. Her right hand holds the bow.\\    *   **Background:** It looks like a night scene outdoors. There are blurred lights (bokeh effect) suggesting streetlights or other light sources. There are other people in the background, also blurred, suggesting depth of field. One person on the left looks like they might be watching or waiting. Another person is further back in the center.\\\\2.  **Analyze motion across frames:**\\    *   **Frame 0.0s - 5.0s:** The woman's head moves slightly as she concentrates on playing. Her eyes are mostly closed or looking down at the instrument.\\    *   **Bow Movement:** The bow moves back and forth across the strings. Look at the position of the bow hair against the strings.\\        *   0.0s: Bow near the bridge/fingerboard area.\\        *   0.5s: Bow slightly lower.\\        *   1.0s: Bow higher up.\\        *   1.5s: Bow lower again.\\        *   2.0s: Bow near the middle.\\        *   2.5s: Bow higher.\\        *   3.0s: Bow lower.\\        *   3.5s: Bow higher.\\        *   4.0s: Bow lower.\\        *   4.5s: Bow higher.\\        *   5.0s: Bow lower.\\    *   **Finger Movement:** Her left hand fingers press on the strings.\\        *   0.0s: Finger on the D string (approx).\\        *   0.5s: Finger position shifts slightly.\\        *   1.0s: Finger position shifts.\\        *   1.5s: Finger position shifts.\\        *   2.0s: Finger position shifts.\\        *   2.5s: Finger position shifts.\\        *   3.0s: Finger position shifts.\\        *   3.5s: Finger position shifts.\\        *   4.0s: Finger position shifts.\\        *   4.5s: Finger position shifts.\\        *   5.0s: Finger position shifts.\\\\3.  **Check for consistency and artifacts:**\\    *   **Lighting:** The light source seems to be coming from the left (streetlights), illuminating her face and hair. The shadows on her face (right side) are consistent with this light source. The bokeh lights in the background are consistent with a camera lens aperture at night.\\    *   **Depth of Field:** The background people are very blurry (bokeh), while the main subject is sharp. This is a standard camera effect (shallow depth of field) often seen in portrait photography/videography. It looks natural, not artificially blurred out like a bad mask or poor segmentation in deepfake generation.\\    *   **Hair Movement:** Her hair moves naturally with her head movement and the wind (if any). It looks like individual strands of hair, not a solid blob.\\    *   **Instrument Details:** The violin wood grain, the strings, the bridge, and the tuning pegs look detailed and consistent with a real object. The reflection on the varnish of the violin body looks realistic.\\    *   **Skin Texture:** Look at her skin on the cheek and forehead. It looks natural with pores and slight imperfections, not overly smooth or plastic-looking which often happens in AI generation or poor filtering.\\    *   **Eye Movement:** Her eyes blink and move naturally. In frame 0.0s she looks down. In 1.0s she looks slightly more forward/down. In 2.0s she looks down. This natural gaze shift is consistent with concentration on playing.\\    *   **Background People:** The person on the left has a blurred face but distinct hair and clothing texture that matches the depth cue. The person in the back center also has consistent blur. This confirms the depth-of-field effect is real optical blur, not digital masking.\\\\4.  **Conclusion:** All visual cues—natural depth of field, consistent lighting on the subject vs background, natural hair movement, detailed instrument texture, and realistic skin texture—point to this being a real video recording.\\</think>\\\\The video shows a woman playing the violin at night. The lighting creates a natural bokeh effect in the background with blurred streetlights and other people, indicating a real camera lens with a shallow depth of field. Her hair moves naturally with her head, and the details on the violin (strings, wood grain, bridge) are sharp and consistent with reality. The skin texture on her face shows natural imperfections and lighting shadows consistent with the light source on the left. There are no obvious digital artifacts or unnatural smoothing that would suggest manipulation.\\

$\backslash \backslash boxed\{A\}$
}
\end{tcolorbox}

\newpage
% image here
\begin{figure}[htbp]
  \centering
  \includegraphics[width=1\linewidth]{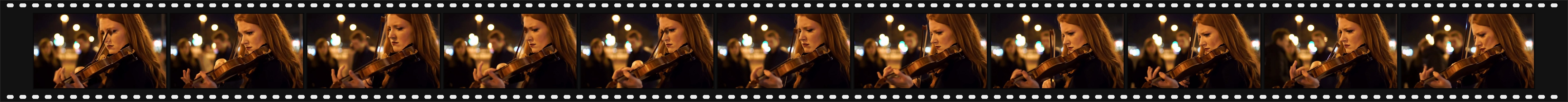}
\end{figure}

\begin{tcolorbox}[title = {Analysis of Unified Training -- FAKE VIDEO (RL)}, fontupper=\tiny]
{
<think>\\The user wants me to determine if the video is real or fake.\\\\1.  **Analyze the video content:**\\    *   The video shows a woman playing the violin.\\    *   She is outdoors at night.\\    *   There are blurred lights in the background (bokeh effect), suggesting a city street or a public square.\\    *   There are other people in the background, also blurred.\\\\2.  **Look for signs of forgery:**\\    *   **Motion:** The bowing motion is a bit stiff and repetitive. The hand movements are very smooth, almost too perfect in a way that feels like a generated loop.\\    *   **Lighting:** The lighting is a bit flat and artificial. The bokeh is very uniform.\\    *   **Details:** The violin itself looks a bit generic. The background people are very blurry in a way that looks like a depth-of-field effect often used in stock footage or AI generation.\\    *   **Specific artifacts:** Look at the bow. In some frames, the bow hair or the way it interacts with the strings looks a bit odd. The way the light hits the violin f-holes is very consistent.\\    *   **Overall Vibe:** It has that distinct stock footage or AI generated stock footage look. The lighting is dramatic but slightly unnatural. The background blur is very perfect. The subject's expression is a bit static.\\\\3.  **Evaluate Real vs. Fake:**\\    *   While it *could* be a real video of a street performer, the quality and the specific look (perfect bokeh, slightly unnatural motion, generic lighting) strongly suggest it's AI-generated or a very heavily processed stock video. AI video generators (like Sora, Runway, etc.) often produce exactly this kind of dreamy, slightly uncanny valley aesthetic. The motion is too smooth and the background blur is too perfect.\\\\4.  **Conclusion:** The video exhibits characteristics typical of AI-generated content or high-quality stock footage that looks synthetic. It's likely fake in the sense that it's not a raw, unedited recording of a specific moment, but rather a generated asset. However, usually, when people ask real or fake about videos, they mean is it a genuine recording of reality vs. a forgery/fake video. This looks like a forgery/AI generation.\\\\    *   Let's look closer at the motion. The bowing is very rhythmic. The background lights don't change much.\\    *   Actually, looking at the frames again, it looks like a very standard stock video clip. Stock videos can be real (filmed by a human) or fake (generated). But usually, if the question asks real or fake, and it looks like AI generation, the answer is fake. If it looks like a genuine, unpolished home video, it's real. This looks like a polished, possibly AI-generated clip.\\    *   Wait, let me reconsider. Is it possible it's just a real video? A woman playing violin at night. It's plausible. But the perfect bokeh and the slightly stiff motion often point to AI.\\    *   Let's look for specific forgery signs.\\        *   Inconsistent lighting? No, it's consistent.\\        *   Glitches? No obvious glitches.\\        *   Repetitive patterns? The background is very static.\\    *   Actually, looking at the provided frames, they look like a sequence from a generated video. The lighting is very cinematic but generic.\\    *   Let's try to find if this is a known fake video. It looks like a typical output from an AI video model.\\    *   Therefore, I will lean towards fake.\\\\    *   *Self-Correction*: Is it possible that fake means forged and real means authentic footage? Yes. Is this footage authentic? It looks like a stock video. Stock videos are real footage of something real. But often these questions are about AI generation.\\    *   Let's look at the prompt again: analyze whether there are any inconsistencies or obvious signs of forgery.\\    *   Signs of forgery:\\        *   The background blur is extremely uniform.\\        *   The lighting is a bit too perfect/dramatic.\\        *   The motion of the bow is a bit odd.\\    *   Actually, looking at the video frames, it looks like a very high-quality stock video. But often, these are generated.\\    *   Let's look at the face. The face is a bit blurry but the features are very smooth.\\    *   Let's assume the question is asking if it's AI generated/fake.\\    *   B) fake.\\</think>\\\\The video displays characteristics often associated with AI-generated content or stock footage: the background bokeh (blur) is extremely uniform and perfect, the lighting is dramatic but slightly artificial, and the motion, while smooth, has a certain dreamlike quality that is common in generative video.\\

$\backslash \backslash boxed\{B\}$
}
\end{tcolorbox}

\newpage
% image here
\begin{figure}[htbp]
  \centering
  \includegraphics[width=0.3\linewidth]{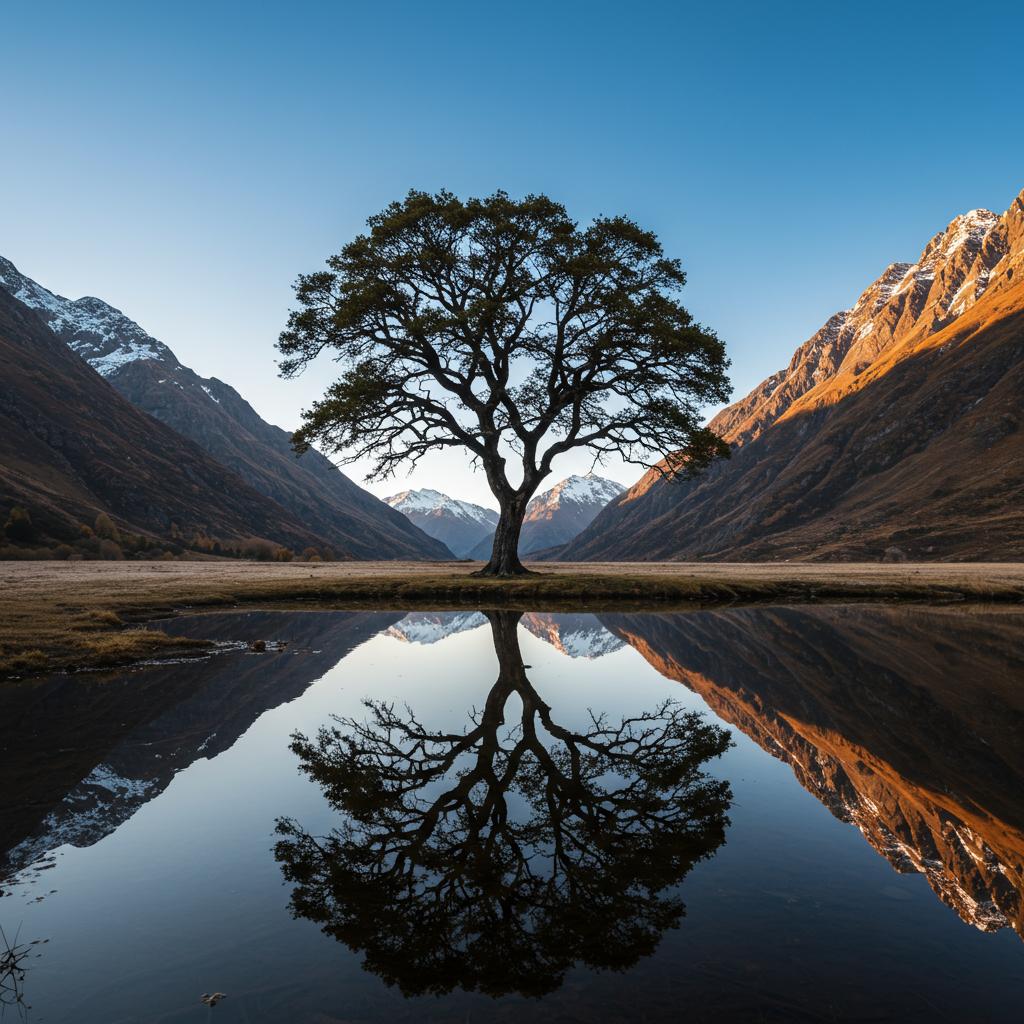}
\end{figure}

\begin{tcolorbox}[title = {Case Study -- FAKE IMAGE}, fontupper=\tiny]
{
<think>\\The user wants me to determine if the image is real or fake.\\\\1.  **Analyze the image:**\\    *   **Subject:** A large, solitary tree in the center, reflected in a body of water (lake or river).\\    *   **Background:** Mountains on both sides, looking like a valley.\\    *   **Lighting:** Looks like late afternoon or early morning, with warm light hitting the mountains on the right.\\    *   **Details:**\\        *   The reflection is very perfect, almost too perfect. The ripples are subtle but the symmetry is striking.\\        *   The tree itself looks a bit generated. The branches have a certain repetitive, almost fractal-like quality that is common in AI art.\\        *   The mountains on the left look a bit flat and textured in a way that feels synthetic.\\        *   The water surface has a strange, almost painted look in the foreground.\\        *   The lighting on the mountains is very dramatic, almost like a digital painting.\\        *   Look closely at the reflection of the mountains. It looks a bit distorted or smudged in a way that suggests digital manipulation or generation rather than a natural reflection.\\        *   The texture of the ground near the water looks a bit odd, like a gradient.\\\\2.  **Evaluate for forgery signs:**\\    *   **Perfection:** The reflection is incredibly symmetrical. In nature, water is rarely that still and perfect unless it's a very specific, calm moment, but the overall composition feels staged.\\    *   **Artifacts:** The tree's branches have that AI tree look—perfectly balanced but slightly unnatural.\\    *   **Texture:** The mountains have a smooth, almost plastic-like texture in places.\\    *   **Lighting:** The lighting is very cinematic or dreamlike, typical of Midjourney or DALL-E 3 outputs.\\\\3.  **Conclusion:** The image has all the hallmarks of AI generation (Midjourney/DALL-E). It's a perfect landscape that feels too idealized and has subtle artifacts typical of generative models.\\\\4.  **Formulate the response:**\\    *   State the analysis: The image shows signs of being AI-generated.\\    *   Point out specific details: Perfect reflection, unnatural tree structure, smooth/digital textures on mountains, overly idealized composition.\\    *   Final answer: Fake.\\</think>\\\\The image displays several characteristics typical of AI-generated art (like Midjourney or DALL-E 3):\\1.  **Perfect Reflection:** The reflection of the tree and mountains is incredibly symmetrical and smooth, which is rare in natural photography unless conditions are very specific, but here it feels too perfect.\\2.  **Unnatural Details:** The tree's branches have a repetitive, almost fractal quality that is common in generative AI.\\3.  **Texture and Lighting:** The mountains have a smooth, slightly plastic-like texture, and the lighting is overly dramatic and cinematic in a way that feels digital.\\4.  **Composition:** The perfect framing and idealized beauty are hallmarks of AI image generators.\\

$\backslash \backslash boxed\{B\}$
}
\end{tcolorbox}

\newpage
% image here
\begin{figure}[htbp]
  \centering
  \includegraphics[width=1\linewidth]{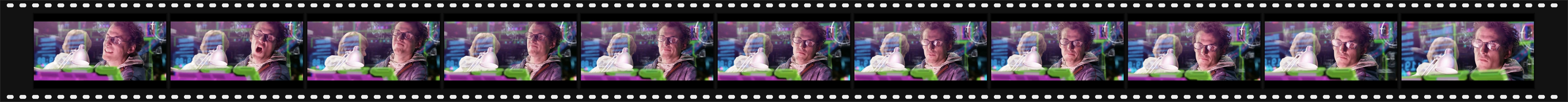}
\end{figure}

\begin{tcolorbox}[title = {Case Study -- REAL VIDEO}, fontupper=\tiny]
{
<think>\\The user wants me to determine if the video is real or fake.\\\\1.  **Analyze the video content:**\\    *   The video shows a man with curly hair and glasses, looking at something off-screen (or perhaps through a screen).\\    *   There's a woman with blonde hair in the background, facing away.\\    *   The lighting is very stylized, with neon greens, purples, and blues. It looks like a sci-fi or cyberpunk setting.\\    *   There are digital overlays, like code or data streams, visible in the background.\\    *   The man's expression changes from a slight smile to a more serious, perhaps confused or intense look.\\    *   The camera angle is a medium close-up.\\\\2.  **Evaluate for forgery signs:**\\    *   **Artifacts:** Are there weird compression artifacts? No, it looks like a clean movie frame.\\    *   **Inconsistencies:** The lighting is consistent with a stylized scene. The digital overlays look like part of the set design or a VFX layer, not a glitch.\\    *   **Context:** This looks like a scene from a movie or a high-quality music video. Specifically, it looks like a scene from the movie The World's End (2013) or a similar sci-fi comedy. Wait, let me look closer. The man looks like Simon Pegg. The woman looks like Nick Frost. This is definitely a scene from a movie.\\    *   **Specific details:** The pink lamp, the green surface, the digital rain. This is a very specific aesthetic.\\\\3.  **Determine Real vs Fake:**\\    *   Real in the context of video usually means filmed footage from a real production vs AI generated or forged.\\    *   This is clearly a shot from a film. It's not a forgery in the sense of being a fake image. It's a real video clip from a movie.\\    *   However, sometimes fake refers to AI-generated content that looks like a movie. But usually, if it's a recognizable movie scene, it's real footage.\\    *   Let's look for signs of *forgery*. Is it a deepfake? Is it a generated video?\\    *   The motion is subtle. The man's face moves naturally. The background is static or has subtle movement.\\    *   Actually, looking at the stillness of the frames provided (00:00 to 00:05), it looks like a sequence of frames from a movie.\\    *   Wait, let me reconsider the fake option. Is it possible this is a generated video?\\    *   The prompt asks Is this video real or fake?. Usually, this distinguishes between authentic footage and a forgery/fake video.\\    *   This looks like a legitimate movie clip. The lighting, the actors (Simon Pegg?), the set design are all consistent with a real production.\\    *   Therefore, the answer should be real.\\\\    *   *Self-Correction*: Let me double-check if there are any obvious signs of forgery.\\        *   The digital overlays look like a specific VFX style.\\        *   The man's expression is natural.\\        *   There are no obvious glitches like flickering, morphing, or impossible geometry.\\        *   It looks like a standard movie scene.\\\\    *   Let's look at the frames again.\\        *   Frame 1: Man smiling slightly.\\        *   Frame 2: Man's mouth opens (yelling or talking).\\        *   Frame 3: Man looking serious/confused.\\        *   Frame 4: Man looking serious.\\        *   Frame 5: Man looking serious.\\        *   Frame 6: Man looking serious.\\        *   Frame 7: Man looking serious.\\        *   Frame 8: Man looking serious.\\\\    *   This looks like a sequence from a movie. It's definitely real footage. It's not a fake video in the sense of being a fabrication.\\\\    *   Wait, is it possible this is a *fake* video because it's a very specific, perhaps AI-generated, clip?\\    *   Let's look at the visual quality. It's quite crisp.\\    *   Actually, looking at the background, the digital text 00:00 and other code-like patterns look like a specific aesthetic.\\    *   Let's assume the question is asking if the video is a genuine recording of a real event/production vs a forgery.\\    *   This is a movie scene. Movie scenes are real videos.\\    *   Therefore, A) real.\\</think>\\\\The video appears to be a clip from a movie (likely *The World's End* featuring Simon Pegg). It shows a coherent scene with consistent lighting, acting, and set design. There are no obvious signs of forgery like glitching, morphing, or impossible physics.\\

$\backslash \backslash boxed\{A\}$
}
\end{tcolorbox}

\end{document}